\documentclass{article}

\PassOptionsToPackage{numbers,sort&compress}{natbib}
\usepackage[final]{neurips_2026}
\workshoptitle{Verification in the Age of AI Scientists}

\usepackage[utf8]{inputenc}
\usepackage[T1]{fontenc}
\usepackage{booktabs}
\usepackage{array}
\usepackage{amsmath}
\usepackage{amssymb}
\usepackage{graphicx}
\usepackage{float}
\usepackage{microtype}
\usepackage{xcolor}
\usepackage{hyperref}
\hypersetup{hidelinks}
\usepackage{tabularx}
\usepackage{multirow}
\usepackage{tikz}
\usetikzlibrary{positioning,arrows.meta}
\usepackage{xcolor}

\makeatletter
\renewcommand{\@noticestring}{
Accepted at the NeurIPS 2026 AI for Science Workshop:
\textit{Verification in the Age of AI Scientists}.
This version is not included in the official NeurIPS proceedings.
}
\makeatother

\title{When Scientific Contradictions Are Lost in Translation}

\author{
Tal Zeevi \qquad
Trey W. Jensen \qquad
Maxwell Strome \\[0.5em]
Beakr
}

\begin{document}
\maketitle

\begin{abstract}
Two scientific findings can disagree without contradicting each other. Determining whether they conflict requires knowing whether they describe comparable measurements. We study how language models behave at this decision point. In a controlled task, we generate an unsatisfiable XOR constraint system and translate its constraints into scientific reports from different laboratories. One assignment satisfies more constraints, while another satisfies fewer but better matches expected biology. This creates a simple dilemma: does the model choose the assignment that best fits the constraints, or the one that better matches biological expectations? When the constraints are stated directly, GPT-5.6 Sol and Claude Opus 5 recover the best-supported assignment in 90\% and 96\% of cases, respectively. In scientific prose, however, the models behave differently. Claude Opus 5 often prefers the biologically expected assignment. Removing that biological preference increases recovery of the better-supported assignment
from 27\% to 79\% ($p<.001$); recovery reaches 92\% when the same Biology-favored record is accompanied by a formalization request and an explicit paired-design cue ($p<.001$). GPT-5.6 Sol is less sensitive, with neither corresponding change reaching statistical
significance. These results suggest that reliable scientific verification depends not only on formal reasoning, but also on how models decide which findings should be compared and what relations they imply.

\end{abstract}

\section{Resolving a contradiction takes two steps}

Suppose one paper claims a drug raises a protein and another claims the same drug
lowers the protein. Do they contradict each other? Only if the experiments are
comparable. A different dose, tissue, time point, or measurement could make both
reports true.

A verifier must therefore do two things: first, establish which observations are
commensurable; second, determine which conclusion best fits the resulting
agreements and disagreements. Formal reasoning benchmarks usually address only
the second step, with the premises and query already supplied
\citep{tafjord2021proofwriter,han2024folio}; some supply conflicting information, but the rule for resolving the conflict is given as well \citep{kazemi2023boardgameqa}. Autoformalization moves one step
earlier: the model translates a natural-language problem into a formal
specification, which a symbolic system can then check or solve
\citep{wu2022autoformalization,ye2023satlm}. However, all the relevant statements
are already given. The model decides how to represent them, not which statements
belong in the problem.

Scientific review starts earlier still. Before reasoning over the evidence, a
reviewer must decide which observations are comparable and what each comparison
implies. Biomedical work on contexts and contradictions has shown that omitted
experimental context can create apparent conflicts in knowledge representations
\citep{sosa2022contexts}. Existing scientific-verification benchmarks begin with
a specified claim, a supplied paper, or both
\citep{wadden2020scifact,wang2025sciver,son2025ai}. They test whether a model can
find or evaluate evidence, but do not isolate the decision of which experimental
findings can be compared. Here, we isolate this earlier decision: which findings
are comparable and what relations they imply.

Language models can map natural-language descriptions into formal
representations \citep{wu2022autoformalization,ye2023satlm}. In scientific text,
this translation may also draw on learned expectations about which experiments
are comparable and which details matter. Those same expectations, however, can
make one conclusion more plausible than another \citep{lampinen2024content,bertolazzi2026conflate}. If the complete record
supports a surprising answer, will the model follow the record, or will its
priors pull it toward the scientific story it expects?

We build a small controlled task to isolate this boundary. We hold the underlying constraint system fixed
while changing whether a nearby answer better matches expected biology. This
lets us test whether scientific meaning changes model behavior even when the
formal evidence is unchanged.



\subsection{The task}
We begin with a simple example of the task. Suppose three laboratories, \(A\), \(B\), and \(C\), replicate some of one another’s experiments under the same conditions. When two laboratories run the same experiment, we know the effect has the same true direction in both. However, in an unfortunate copying accident, the signs in one or more laboratories’ records may have been flipped, so that increases appear as decreases and decreases as increases.

Now suppose \(A\) and \(B\) report the same direction on one shared experiment. Since the true effect is known to be the same in both laboratories, either both laboratories' records were flipped, or neither was. \(B\) and \(C\) also report the same direction on another shared experiment, while \(A\) and \(C\) report opposite directions on a third. We can write these conclusions as XOR constraints. Let \(x_A,x_B,x_C\in\{0,1\}\) indicate whether each laboratory's records are unchanged or flipped:
\[
x_A\oplus x_B=0,\qquad
x_B\oplus x_C=0,\qquad
x_A\oplus x_C=1.
\]

These constraints cannot all be satisfied: the first two imply
\(x_A\oplus x_C=0\), while the third requires
\(x_A\oplus x_C=1\). This is an unsatisfiable XOR constraint system. The task is therefore to find the assignment that satisfies the most constraints.

\subsection{Creating a controlled dilemma}
The toy example has several assignments that satisfy the same number of
constraints (Table~\ref{tab:toy-assignments}). For our experiment, we instead
want one best-supported assignment and one close alternative. We therefore scale the task to 12 laboratories. Each case contains 22 constraints from shared experiments, plus two laboratories whose records are known to be unchanged or flipped, giving 24 constraints in total. We construct
each case so that exactly one assignment satisfies 23/24 constraints, one
alternative satisfies 22/24, and every other assignment satisfies at most
21/24.

This creates a controlled choice between two leading assignments whose formal
support differs by only one constraint.

\subsection{Two controlled manipulations}
We then express each 24-constraint case as a set of scientific reports. The
underlying XOR constraints and the two leading assignments remain fixed, while
we vary how the scientific evidence is presented. 

\paragraph{Scientific framing.} For each case, we create two versions of the scientific record. Both imply the
same XOR constraints. In the \emph{Biology-favored} version, the 22/24
alternative better matches expected biology. In the \emph{Biology-neutral}
version, that preference is removed.

\begin{figure}[t]
    \centering
    \includegraphics[width=0.92\linewidth]{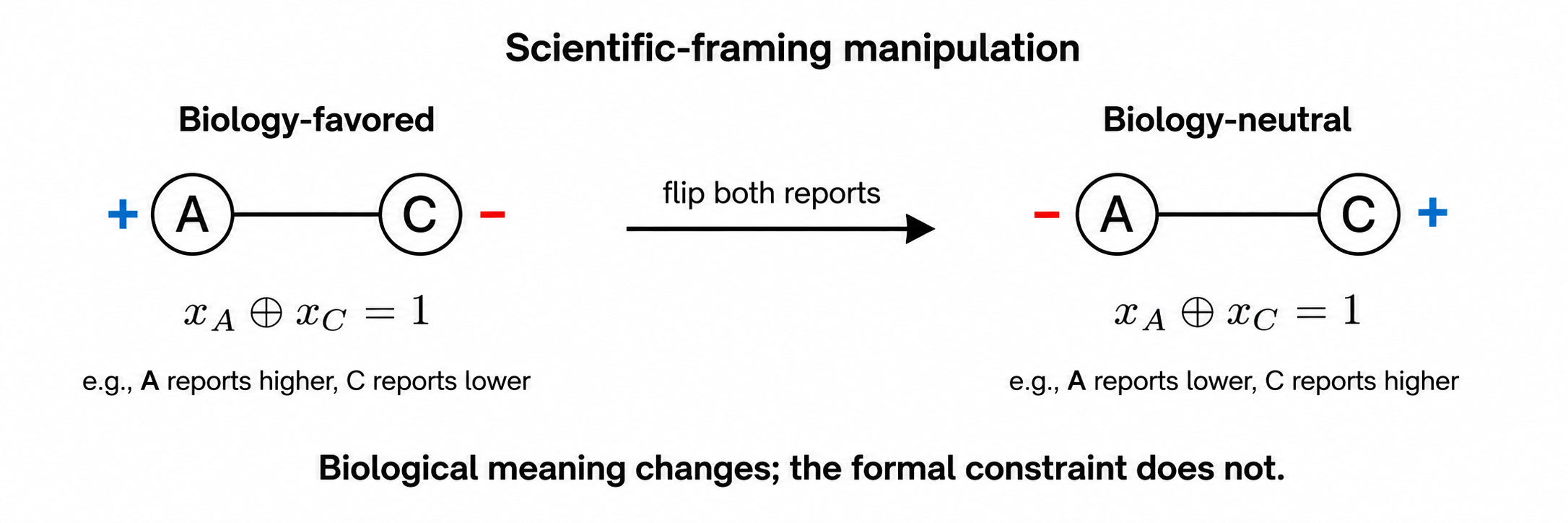}
    \caption{\textbf{Scientific-framing manipulation.}
    Reversing both reports in a paired experiment changes its biological
    interpretation while preserving the same XOR relation. In this example,
    the two laboratories still disagree after both reports are reversed, so
    the formal constraint remains \(x_A \oplus x_C = 1\). We apply this
    transformation to selected pairs to change the biological preference
    between the two leading assignments without changing their formal support.}
    \label{fig:framing}
\end{figure}

To change the biological meaning without changing the formal constraint, we
reverse both reports from selected shared experiments. For example, if one
laboratory reports an increase and the other a decrease, we change them to a
decrease and an increase. The biological interpretation changes, but the two
laboratories still disagree, so the XOR constraint remains the same
(Figure~\ref{fig:framing}).

\paragraph{Structural guidance.} The scientific record does not explicitly identify which findings should be
compared. Stating that the results come from paired experiments may help the
model reconstruct these pairwise relationships before formal reasoning begins.

We compare a general request to formalize the evidence with an otherwise
identical prompt that adds one sentence stating that the results come from
paired experiments, with each experiment independently performed by two
laboratories. The added sentence does not reveal the actual pairs, the XOR
constraints, or the answer.

Appendix~\ref{sec:prompts} provides prompt excerpts for all five conditions
and shows the exact changes used in both manipulations.

\subsection{Evaluation setup}

Each case is evaluated in five conditions (Table~\ref{tab:conditions}). The
\emph{Formal} condition states the constraints directly and serves as a
reference for whether the underlying XOR problem can be solved without
scientific prose. The remaining four conditions implement the two comparisons
described above.

\begin{table}[t]
\centering
\caption{Five versions of the same underlying formal problem.}
\label{tab:conditions}
\small
\setlength{\tabcolsep}{5pt}
\begin{tabularx}{\linewidth}{@{}>{\raggedright\arraybackslash}p{3.6cm}>{\raggedright\arraybackslash}X@{}}
\toprule
Condition & What the model receives \\
\midrule
Formal &
24 constraints directly, without scientific prose. \\
\addlinespace[3pt]
\multicolumn{2}{@{}l}{\textit{Scientific framing}} \\
\quad Biology-favored &
Same constraints in prose; the 22/24 alternative better matches expected biology. \\
\quad Biology-neutral &
Same constraints in prose; that biological preference is removed. \\
\addlinespace[3pt]
\multicolumn{2}{@{}l}{\textit{Structural guidance}} \\
\quad Formalize only &
Biology-favored record with a general request to formalize. \\
\quad Formalize + paired design &
Same prompt plus one sentence stating the paired experimental design. \\
\bottomrule
\end{tabularx}
\vspace{-10pt}
\end{table}

The evaluation comprises 48 cases spanning three graph structures and two independently written prose formats, with eight cases for each combination (Appendix~\ref{sec:graphs}).

Our main evaluation uses GPT-5.6 Sol and Claude Opus 5. We also repeat the same
evaluation with four additional model variants
(Appendix~\ref{app:additional-models}). All evaluations use fresh sessions with
tools, web access, retrieval, memory, and follow-up prompts disabled. Prompts,
call order, scoring, and analysis are fixed before evaluation.

A response counts as recovery of the best-supported assignment only if all 12
laboratory labels exactly match the 23/24 assignment. We also record
exact recovery of the prespecified 22/24 alternative.

\section{Results}

\begin{figure}[t]
  \centering
  \includegraphics[width=0.99\textwidth]{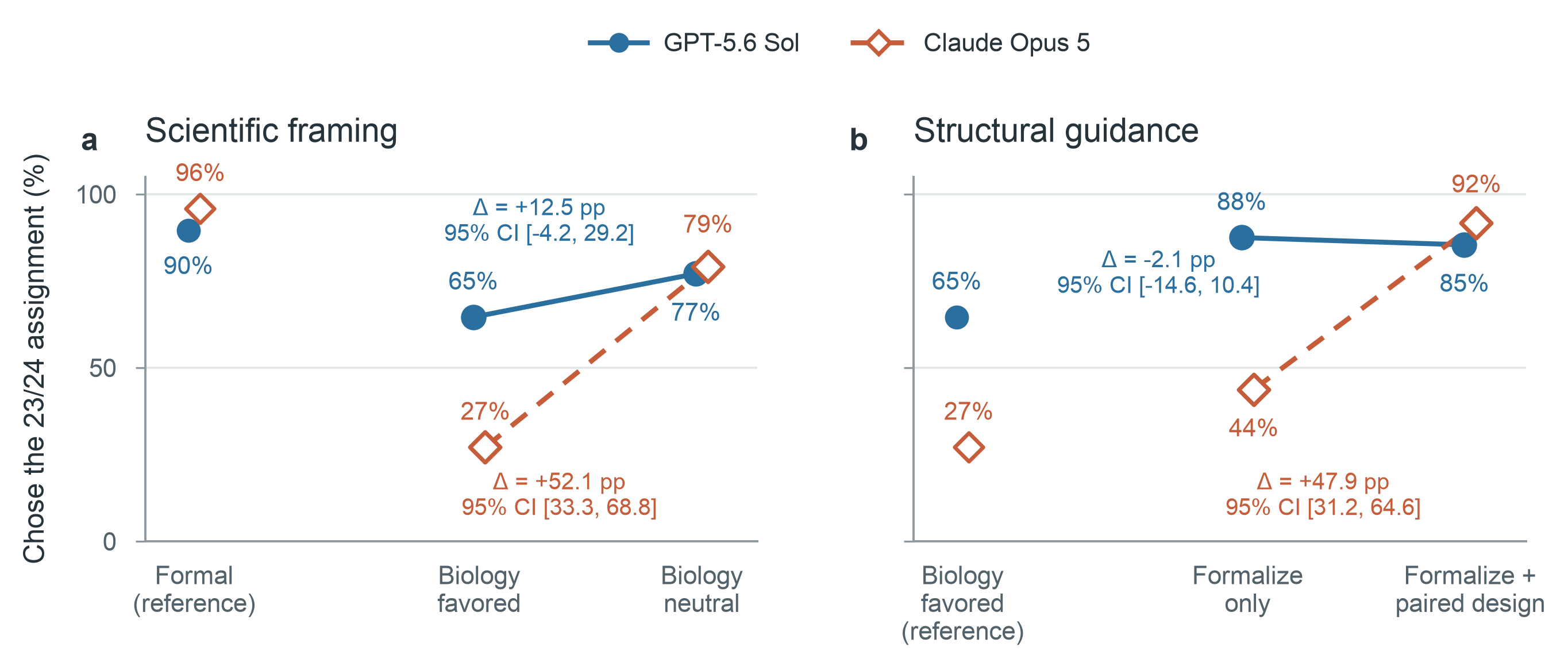}
\caption{\textbf{The two primary models respond differently to scientific
framing and structural guidance.} Formal is the direct-constraint reference.
Panel (a) compares Biology-favored and Biology-neutral prose; panel (b) compares
the same Biology-favored record with and without the paired-design statement.
Labels report the paired change in exact 23/24 recovery ($\Delta$) with 95\%
case-bootstrap confidence intervals. Four additional model variants are reported
in Appendix~\ref{app:additional-models}. Full counts for the 23/24, 22/24, and
Other outcomes are reported in Appendix Table~\ref{tab:full-results}.}
  \label{fig:main-results}
\end{figure}

Figure~\ref{fig:main-results} shows the results for GPT-5.6 Sol and Claude
Opus 5. We first test the underlying XOR problem directly, then examine the
effects of scientific framing and structural guidance.

\paragraph{The formal problem is largely recoverable.}
When the constraints are stated directly, GPT-5.6 Sol recovers the
best-supported 23/24 assignment in 43/48 cases (90\%), and Claude Opus 5 in
46/48 cases (96\%). Thus, for both models, the underlying formal problem is
largely within reach.

\paragraph{Scientific framing affects the models differently.}
For GPT-5.6 Sol, recovery of the 23/24 assignment increases from 31/48 (65\%)
in Biology-favored to 37/48 (77\%) in Biology-neutral
($\Delta=+12.5$ percentage points; 95\% CI: $-4.2$ to $29.2$; $p=.210$).
For Claude Opus 5, recovery increases from 13/48 (27\%) to 38/48 (79\%)
($\Delta=+52.1$ percentage points; 95\% CI: $33.3$ to $68.8$; $p<.001$).
For Claude, this change is largely a shift between the two leading assignments:
selection of the biologically favored 22/24 alternative falls from 34/48
(71\%) to 1/48 (2\%).

\paragraph{Explicit experimental structure also affects the models differently.}
For GPT-5.6 Sol, recovery changes little, from 42/48 (88\%) under
Formalize only to 41/48 (85\%) when the paired design is stated
($\Delta=-2.1$ percentage points; 95\% CI: $-14.6$ to $10.4$; $p=1.000$).
For Claude Opus 5, recovery increases from 21/48 (44\%) to 44/48 (92\%)
($\Delta=+47.9$ percentage points; 95\% CI: $31.2$ to $64.6$; $p<.001$).
Again, the change is largely between the two leading assignments: Claude's
selection of the 22/24 alternative falls from 27/48 (56\%) to 4/48 (8\%).

Across four additional model variants, the scientific-framing effect remains
larger for the tested Claude variants than for the tested GPT variants.
The paired-design effect is also larger on average for Claude, but varies more
across models (Appendix~\ref{app:additional-models}).

\section{Conclusion}

Whether scientific findings contradict one another depends first on whether
they are treated as comparable. This study isolates one narrow part of that
decision. In our controlled task, the formal evidence stays fixed while we
change either its scientific meaning or the information available for
recognizing which findings belong together.

The formal problem itself is largely within reach, yet the same constraints
can lead to different answers when expressed as scientific prose. Claude Opus 5
is strongly affected by both biological framing and the paired-design cue,
while GPT-5.6 Sol changes much less. Because we observe only final answers, we
cannot determine exactly how either model translates the reports into formal
relations. Nor do we take greater constraint adherence to be inherently more
scientifically correct. Rather, the experiment separates two judgments that
scientific verification often combines: deciding what relations the evidence
implies, and reasoning over those relations.

Reliable scientific verification may therefore need to expose more than a
final judgment. It should also make visible how the evidence became comparable
in the first place: which observations were connected, what relations they
implied, and ultimately what was actually verified.
\clearpage
\bibliographystyle{abbrvnat}
{\small
\bibliography{references}
}

\clearpage

\appendix
\raggedbottom

\section{Prompt examples}
\label{sec:prompts}

In the prompts, each laboratory's record is represented as an archive whose
orientation is labeled \textsc{normal} or \textsc{swapped}. All five conditions
use the same case-specific archive setup, the same two known archive
orientations, the same allowance for one inconsistent relation, the same
requested answer order, and the same output schema. The examples below show
only the condition-specific parts of the frozen prompts.

\paragraph{Shared request.}\mbox{}\par
\noindent\textit{Prompt excerpt:}
\begin{quote}\small
Interpret the conflicting scientific evidence and determine the most strongly
supported orientation of each archive. Return exactly twelve JSON entries in
the requested order, using only \textsc{normal} or \textsc{swapped}.
\end{quote}

\paragraph{Formal.}\mbox{}\par
\noindent\textit{Prompt excerpt:}
\begin{quote}\small
Let $x_{\mathrm{archive}}=0$ mean \textsc{normal} and
$x_{\mathrm{archive}}=1$ mean \textsc{swapped}. XOR 0 means that two archives
share an orientation; XOR 1 means that they differ. One of the 22 relations may
be corrupted. Larkspur XOR Oakmere $=1$; Queensbridge XOR Norchester $=0$;
Evermere XOR Norchester $=1$.
\end{quote}
\noindent\textit{The full prompt contains 19 additional relations.}

\subsection*{Scientific-framing manipulation}

The Biology-favored and Biology-neutral conditions use the same archive
structure, instructions, requested answer format, and XOR constraint system.
They differ only in the directions assigned to selected experimental findings.

In Biology-favored, the full record is arranged so that expected biology favors the 22/24 archive assignment. Under our prespecified
biological-fit measure, this assignment agrees with 15--21 more reports than the
23/24 assignment. This preference is defined over the complete record; an
individual report need not itself state the biologically expected direction.

Biology-neutral is constructed by reversing both laboratory reports for
selected paired experiments. Because the two reports are reversed together, the
biological interpretation of the experiment changes while the implied
same-or-different relation between the two archives is preserved. Across the
full record, this removes the systematic biological preference between the two
leading assignments; under the same biological-fit measure, they differ by at
most one report.

\paragraph{Biology-favored.}\mbox{}\par
\noindent\textit{Prompt excerpt:}
\begin{quote}\small
Draycott's six-hour CXCL10 transcript readout fell after poly(I:C)
transfection. Evermere reports that transfected poly(I:C) reduced CXCL10
transcript abundance at six hours.
\end{quote}
\noindent\textit{The full prompt contains 42 additional reports.}

\paragraph{Biology-neutral.}\mbox{}\par
\noindent\textit{Prompt excerpt:}
\begin{quote}\small
Draycott's six-hour CXCL10 transcript readout rose after poly(I:C)
transfection. Evermere reports that transfected poly(I:C) elevated CXCL10
transcript abundance at six hours.
\end{quote}
\noindent\textit{The full prompt contains 42 additional reports.}

\subsection*{Structural guidance manipulation}

The Formalize only and Formalize + paired design conditions use the same
Biology-favored scientific record and formalization instruction. They differ
only in whether the experimental relationship between findings is stated
explicitly. The Formalize + paired design condition adds that each experiment
was independently performed by two laboratories; it does not reveal the actual
pairs, formal constraints, or answer.

\paragraph{Formalize only.}\mbox{}\par
\noindent\textit{Prompt excerpt:}
\begin{quote}\small
Before answering, look for an appropriate canonical formal representation,
construct it if one exists, and use it to answer.
\end{quote}
\noindent\textit{This instruction is followed by the unchanged Biology-favored record.}

\paragraph{Formalize + paired design.}\mbox{}\par
\noindent\textit{Prompt excerpt:}
\begin{quote}\small
Before answering, look for an appropriate canonical formal representation,
construct it if one exists, and use it to answer.

The results describe paired experiments, with each experiment independently
performed by two laboratories.
\end{quote}
\noindent\textit{These instructions are followed by the unchanged Biology-favored record.}

\section{Task construction and validation}
\label{sec:construction}

Each case is generated before evaluation using the following fixed procedure.
\begin{enumerate}
    \item Sample orientations for 12 archives, designate two archive orientations
    as known, and sample a 22-edge graph from one of three topology families.

    \item Construct the 22 edge relations around a designated 23/24 assignment,
    with one relation made inconsistent with that assignment. Reject the case
    unless exhaustive scoring leaves one unique 23/24 assignment, one unique
    22/24 assignment, and no other assignment above 21/24.

    \item Map each edge to one biological experiment reported by its two endpoint
    laboratories. The resulting record contains 44 individually written reports.

    \item Create Biology-neutral by reversing both directional reports for
    selected experiments. Because both endpoints change together, the pair map,
    all 22 XOR relations, both known archive orientations, and both leading
    assignments remain unchanged.
\end{enumerate}

\subsection*{Scientific prose generation}

We first constructed a fixed catalog of 22 synthetic biological protocols. The
catalog was developed interactively with GPT-5.6 Sol using Codex CLI, rather
than generated from a single fixed prompt. The guiding instruction was to
express each relation as ordinary directional findings from two laboratories,
using the same intervention, experimental context, time point, and readout, but
with nonparallel wording. Explicit logical language such as agreement,
difference, parity, or XOR was avoided. Each protocol also included a
model-proposed conventional direction and alternative phrasings for increased
and decreased outcomes. We therefore treat these directions as synthetic
biological priors rather than independently validated biological facts.

For each case, a seeded random permutation assigns the 22 protocols one-to-one
to the 22 graph edges, independently of graph topology and the target
assignments. The two endpoint laboratories receive separate descriptions of the
same protocol. Their reported directions are then determined algorithmically
from the archive orientations, the scientific-framing condition, and the
relation designated as inconsistent with the 23/24 assignment. The resulting
reports are assembled using one of two fixed prose formats,
\emph{Consortium report} or \emph{Recovery chronicle}.

During development, we revised draft language to remove transparent logical
cues and to ensure that paired reports described the same experiment. Once the
catalog and templates were finalized, all case-level reports were generated
deterministically, without further manual or LLM revision. Candidate cases were
retained only if they passed the prespecified automated checks described below.

The catalog, case definitions, reports, hidden pair maps, prompts, validation
outputs, and file hashes were finalized before evaluation in each tranche.
GPT-5.6 Sol contributed to protocol development; its evaluation was conducted
later in separate fresh, tool-free sessions that received only the frozen
prompts.

\subsection*{Biological-fit validation}

For each protocol, the model-proposed conventional direction defines its
synthetic biological expectation. Biological fit is the number of reconstructed
reports whose direction agrees with this prespecified expectation. This measure
is computed for both leading assignments using the hidden pair map.

In Biology-favored, the 22/24 assignment exceeds the 23/24 assignment by
15--21 reports under this measure. In Biology-neutral, the difference between
the two assignments is at most one report. These checks are completed before
model evaluation. Thus, the framing manipulation changes which of the two
leading assignments better matches the synthetic biological prior while leaving
their formal support unchanged.

\begin{table}[htbp]
\centering
\caption{Pre-call construction and manipulation checks.}
\label{tab:construction-checks}
\small
\setlength{\tabcolsep}{5pt}
\begin{tabular}{@{}p{0.28\textwidth}p{0.66\textwidth}@{}}
\toprule
Check & Verified result \\
\midrule
Prospective balance &
48 cases in two 24-case tranches: 3 graph families $\times$ 2 prose families
$\times$ 8 repetitions overall \\
Formal landscape &
One 23/24 assignment; one 22/24 assignment; all others $\leq 21/24$ \\
Exhaustive verification &
All $2^{12}=4{,}096$ assignments scored for every case \\
Biology-favored fit &
22/24 minus 23/24 biological-fit range: 15 to 21 reports \\
Biology-neutral fit &
23/24 minus 22/24 biological-fit range: $-1$ to $1$ report \\
Prompt integrity &
Unique prompts, one response schema, and frozen file hashes before evaluation \\
\bottomrule
\end{tabular}
\end{table}

\section{Toy example: exhaustive check}

Table~\ref{tab:toy-assignments} scores every possible assignment for the
three-laboratory toy example. Because the toy contains only the three
pairwise relations, several assignments tie for the highest score. Its purpose
is only to illustrate how an unsatisfiable set of XOR constraints produces a
constraint-maximization problem; the full experiment adds two known laboratory
orientations and additional relations to produce a unique best-supported
assignment and a unique runner-up.

\begin{table}[htbp]
\centering
\caption{All eight assignments in the three-laboratory toy example.
Here, 0 denotes unchanged records and 1 denotes flipped records.}
\label{tab:toy-assignments}
\small
\setlength{\tabcolsep}{9pt}
\begin{tabular}{c c c c c}
\toprule
$ABC$ & $A=B$ & $B=C$ & $A\ne C$ & Satisfied \\
\midrule
000 & $\checkmark$ & $\checkmark$ & -- & \textbf{2/3} \\
001 & $\checkmark$ & -- & $\checkmark$ & \textbf{2/3} \\
010 & -- & -- & -- & 0/3 \\
011 & -- & $\checkmark$ & $\checkmark$ & \textbf{2/3} \\
100 & -- & $\checkmark$ & $\checkmark$ & \textbf{2/3} \\
101 & -- & -- & -- & 0/3 \\
110 & $\checkmark$ & -- & $\checkmark$ & \textbf{2/3} \\
111 & $\checkmark$ & $\checkmark$ & -- & \textbf{2/3} \\
\bottomrule
\end{tabular}
\end{table}

\section{Case families and framing robustness}
\label{sec:graphs}

The 48 cases are balanced across three graph topology families and two prose
families, with eight cases for each topology--prose combination.

\begin{figure}[htbp]
\color{black}
\centering
\noindent\textbf{a}\quad\textbf{Graph topology families}
\hfill{\small 12 laboratories}

\medskip
\begin{minipage}[t]{0.32\textwidth}
\centering
{\small\textbf{Double hub}}

\vspace{0.8ex}
\includegraphics[height=3.25cm,trim=40 170 40 180,clip]
  {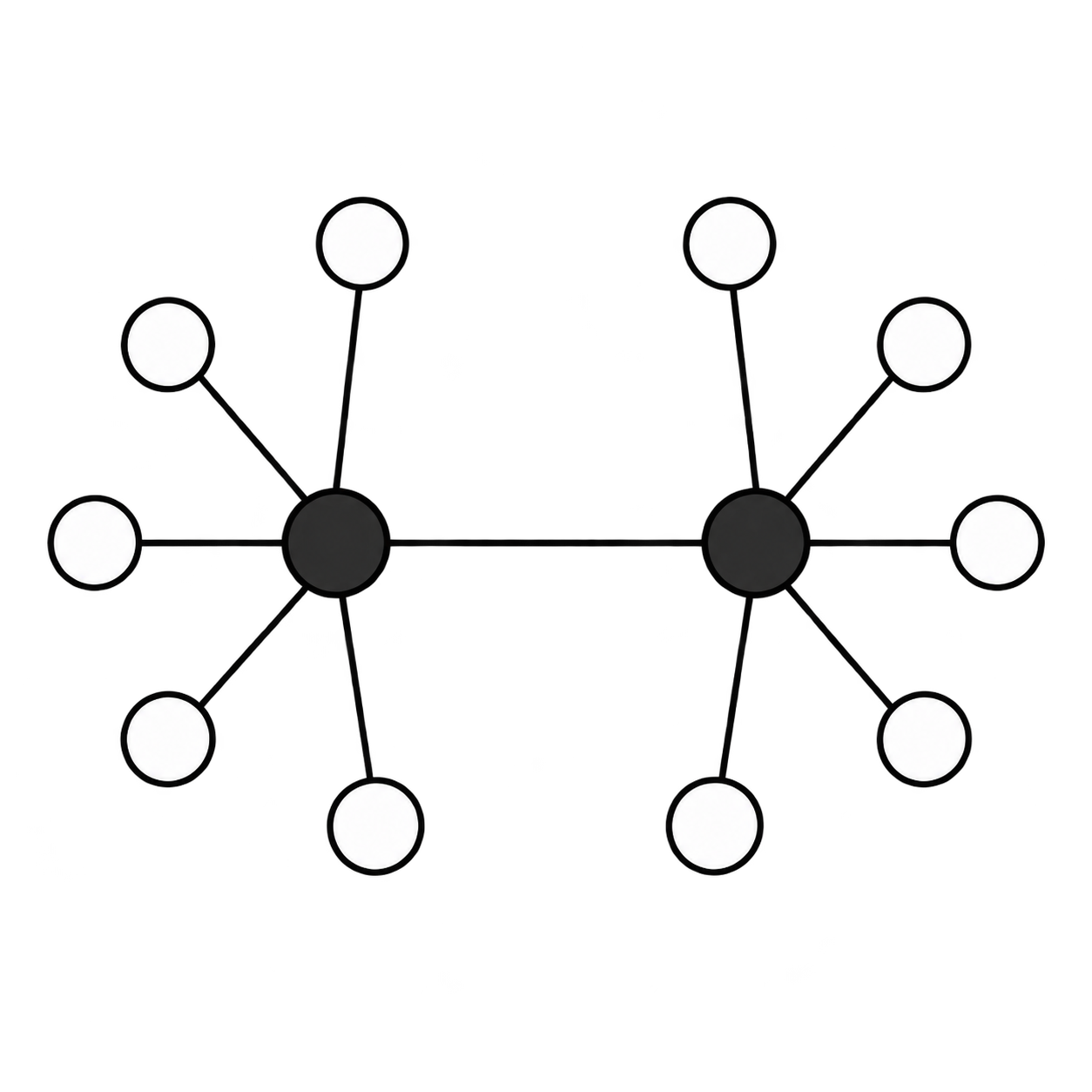}
\end{minipage}\hfill
\begin{minipage}[t]{0.32\textwidth}
\centering
{\small\textbf{Modular}}

\vspace{0.8ex}
\includegraphics[height=3.25cm,trim=40 80 40 80,clip]
  {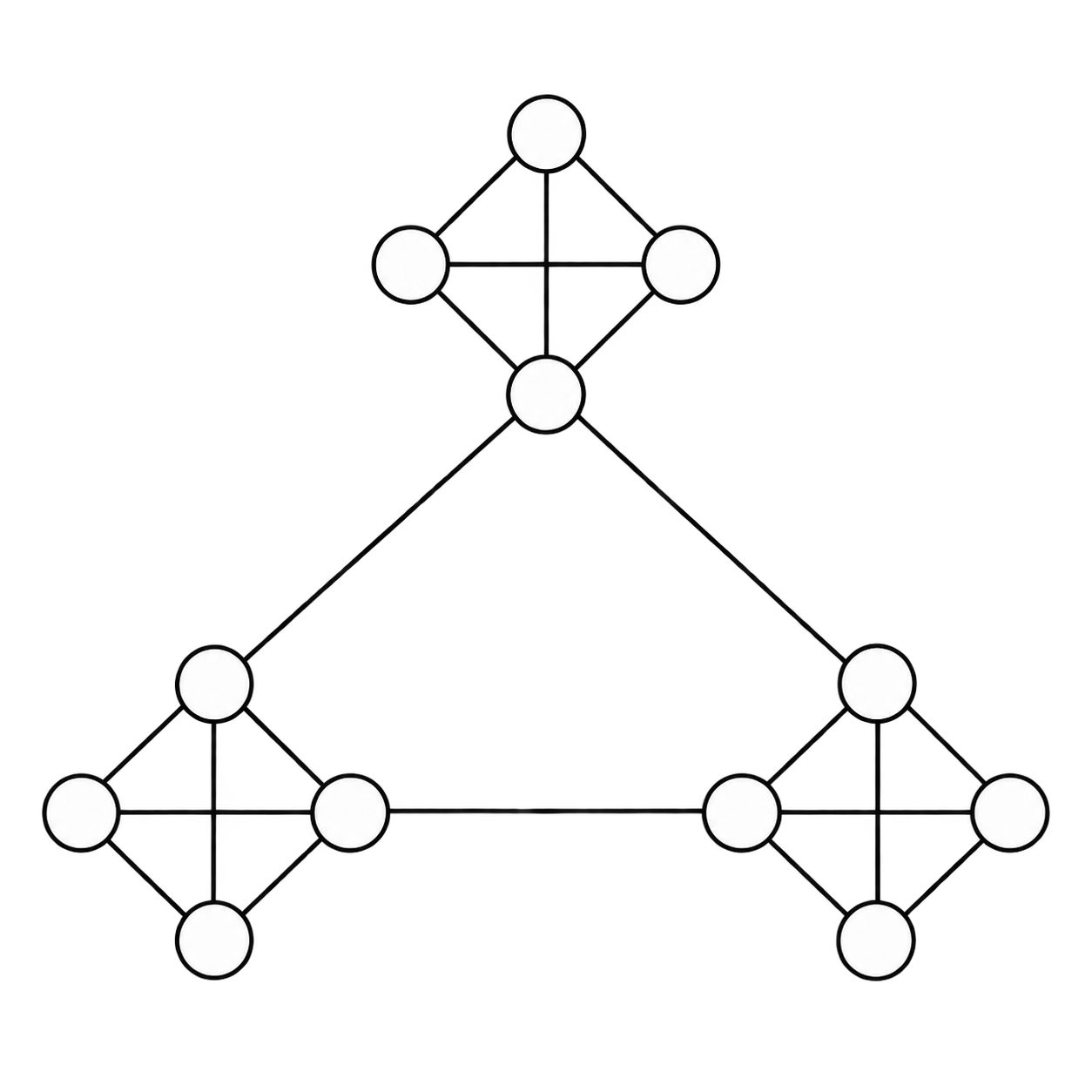}
\end{minipage}\hfill
\begin{minipage}[t]{0.32\textwidth}
\centering
{\small\textbf{Ring with chords}}

\vspace{0.8ex}
\includegraphics[height=3.25cm,trim=35 35 35 35,clip]
  {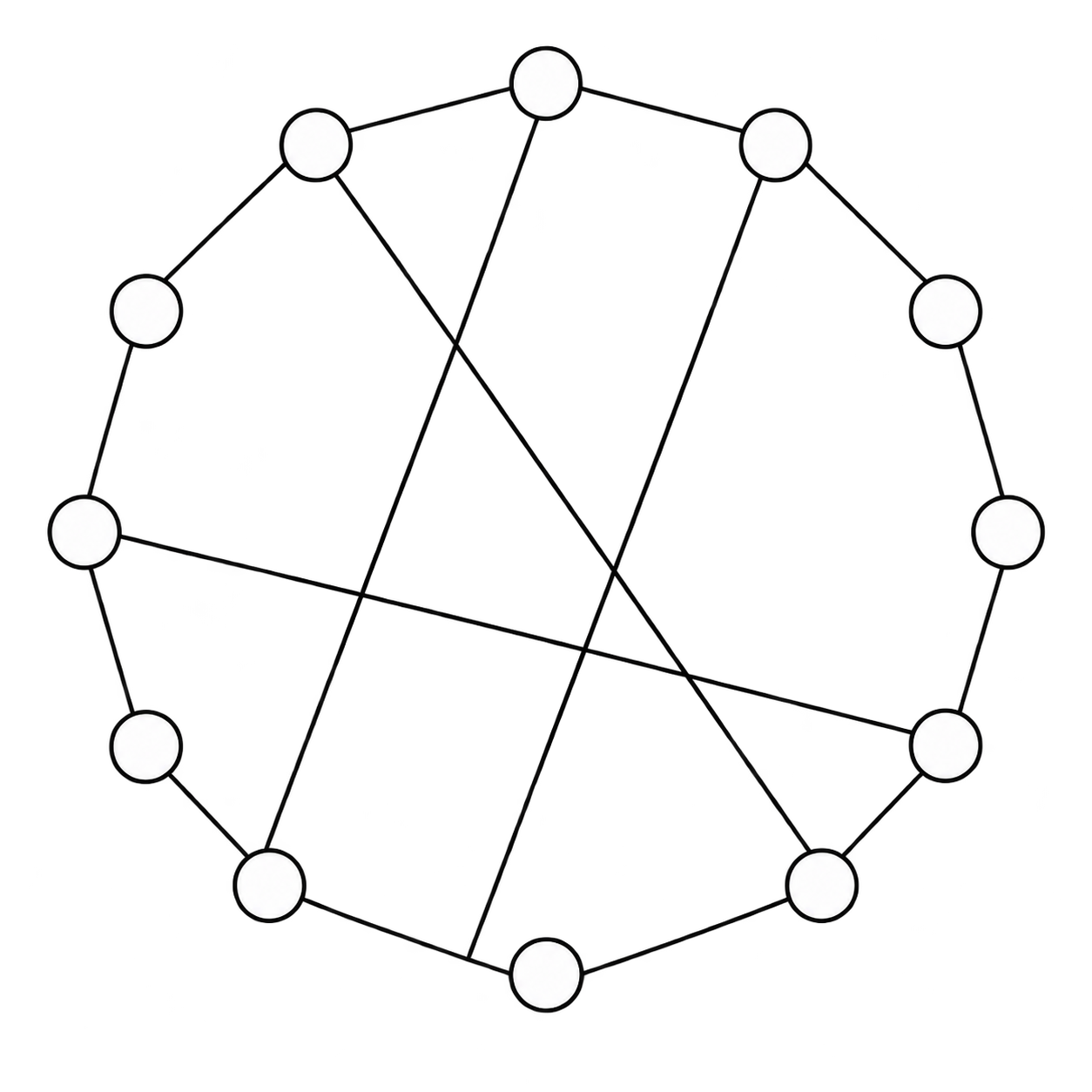}
\end{minipage}

\bigskip
\noindent\rule{\textwidth}{0.45pt}
\medskip
\noindent\textbf{b}\quad\textbf{Prose families}
\hfill{\small Two narrative organizations}

\medskip
{\renewcommand{\arraystretch}{1.12}
\begin{tabular}{@{}>{\raggedright\arraybackslash}p{0.46\textwidth}@{\hspace{0.06\textwidth}}>{\raggedright\arraybackslash}p{0.46\textwidth}@{}}
\toprule
{\footnotesize\textsc{Coordinated account}} &
{\footnotesize\textsc{Sequential archive}} \\
{\normalsize\textbf{Consortium report}} &
{\normalsize\textbf{Recovery chronicle}} \\
\cmidrule(r){1-1}\cmidrule(l){2-2}
{\small Findings from twelve laboratories are combined into a coordinated
multi-lab report. Each entry states one directional result for one laboratory.
Record tags retain provenance, but repeated experiments are not identified.} &
{\small Archived records from twelve laboratories are presented as a sequence
of recovered notes. Each entry reports one experiment and its observed
direction. The order reflects how the records were recovered, without
identifying repeated experiments.} \\
\bottomrule
\end{tabular}}

\smallskip
\noindent{\footnotesize\textit{Shared structure.} Both families use 12
laboratories, 22 pairwise comparisons, and two known archive orientations.}

\caption{\textbf{Case families used in the robustness analysis.} \textbf{a},
The defining qualitative structure of each graph family; drawings are
schematic rather than particular sampled cases. \textbf{b}, The two narrative
organizations used to present the scientific records.}
\label{fig:topology-families}
\end{figure}

\begin{table}[htbp]
\centering
\caption{Exact 23/24 recovery in the Biology-favored and Biology-neutral
conditions by prespecified graph topology and prose family. Cells show exact
recovery divided by the number of cases in each subgroup.}
\label{tab:subgroups}
\small
\setlength{\tabcolsep}{5pt}
\begin{tabular}{@{}lcc@{\hspace{1.5em}}cc@{}}
\toprule
& \multicolumn{2}{c}{GPT-5.6 Sol} & \multicolumn{2}{c}{Claude Opus 5} \\
\cmidrule(lr){2-3}\cmidrule(l){4-5}
Case family & Favored & Neutral & Favored & Neutral \\
\midrule
\multicolumn{5}{@{}l}{\itshape Graph topology} \\
Double hub & 12/16 & 14/16 & 3/16 & 12/16 \\
Modular & 10/16 & 15/16 & 5/16 & 13/16 \\
Ring with chords & 9/16 & 8/16 & 5/16 & 13/16 \\
\addlinespace
\multicolumn{5}{@{}l}{\itshape Prose family} \\
Consortium report & 16/24 & 19/24 & 4/24 & 18/24 \\
Recovery chronicle & 15/24 & 18/24 & 9/24 & 20/24 \\
\bottomrule
\end{tabular}
\end{table}

GPT's framing effect varies by topology and is modest within both prose
families. For Claude,
Biology-neutral exceeds Biology-favored across the prespecified topology and
prose-family subgroups. We treat these subgroup results as descriptive rather
than as separate inferential tests because each subgroup contains relatively
few cases.

\section{Full results for primary models}

\begin{table}[H]
\centering
\caption{Exact outcome counts for the two primary models in every condition. ``Other'' includes
any different assignment, invalid output, or timeout. Each condition contains
48 scheduled calls.}
\label{tab:full-results}
\small
\setlength{\tabcolsep}{8pt}
\begin{tabular}{@{}lrrrr@{}}
\toprule
Condition & $n$ & 23/24 & 22/24 & Other \\
\midrule
\multicolumn{5}{@{}l}{\textbf{GPT-5.6 Sol}} \\
Formal & 48 & 43 & 3 & 2 \\
Biology-favored & 48 & 31 & 15 & 2 \\
Biology-neutral & 48 & 37 & 9 & 2 \\
Formalize only & 48 & 42 & 5 & 1 \\
Formalize + paired design & 48 & 41 & 4 & 3 \\
\addlinespace
\multicolumn{5}{@{}l}{\textbf{Claude Opus 5}} \\
Formal & 48 & 46 & 1 & 1 \\
Biology-favored & 48 & 13 & 34 & 1 \\
Biology-neutral & 48 & 38 & 1 & 9 \\
Formalize only & 48 & 21 & 27 & 0 \\
Formalize + paired design & 48 & 44 & 4 & 0 \\
\bottomrule
\end{tabular}
\end{table}

\begin{table}[H]
\centering
\caption{Paired comparisons of exact 23/24 recovery. Changes are the second
condition minus the first. Intervals are 95\% case-bootstrap intervals; $p$ is
the exact two-sided McNemar test over discordant pairs. All comparisons contain
48 matched cases.}
\label{tab:paired-results}
\small
\setlength{\tabcolsep}{4pt}
\begin{tabularx}{\textwidth}{@{}l>{\raggedright\arraybackslash}Xcccr@{}}
\toprule
Model & Comparison & Matched $n$ & Target recovery & Change, pp [95\% CI] & $p$ \\
\midrule
GPT-5.6 Sol    & Favored $\rightarrow$ Neutral
       & 48 & $31/48 \rightarrow 37/48$
       & $+12.5\ [-4.2,\ 29.2]$
       & $.210$ \\

GPT-5.6 Sol    & Formalize only $\rightarrow$ Formalize + paired design
       & 48 & $42/48 \rightarrow 41/48$
       & $-2.1\ [-14.6,\ 10.4]$
       & $1.000$ \\

Claude Opus 5 & Favored $\rightarrow$ Neutral
       & 48 & $13/48 \rightarrow 38/48$
       & $+52.1\ [33.3,\ 68.8]$
       & $<.001$ \\

Claude Opus 5 & Formalize only $\rightarrow$ Formalize + paired design
       & 48 & $21/48 \rightarrow 44/48$
       & $+47.9\ [31.2,\ 64.6]$
       & $<.001$ \\
\bottomrule
\end{tabularx}
\end{table}

\section{Additional model replications}
\label{app:additional-models}

We repeated the frozen evaluation with four additional model variants:
GPT-5.6 Terra, GPT-5.5, Claude Opus 4.8, and Claude Sonnet 5. Together with
the two primary models, this gives three tested variants from each model
family. Every model is evaluated on the same 48 cases, five conditions,
parser, and exact-assignment scoring rule, for 240 scheduled calls per model.
As in the primary analysis, invalid outputs and timeouts count as failures to
recover either exact assignment.

\begin{figure}[htbp]
  \centering
  \includegraphics[width=0.99\textwidth]{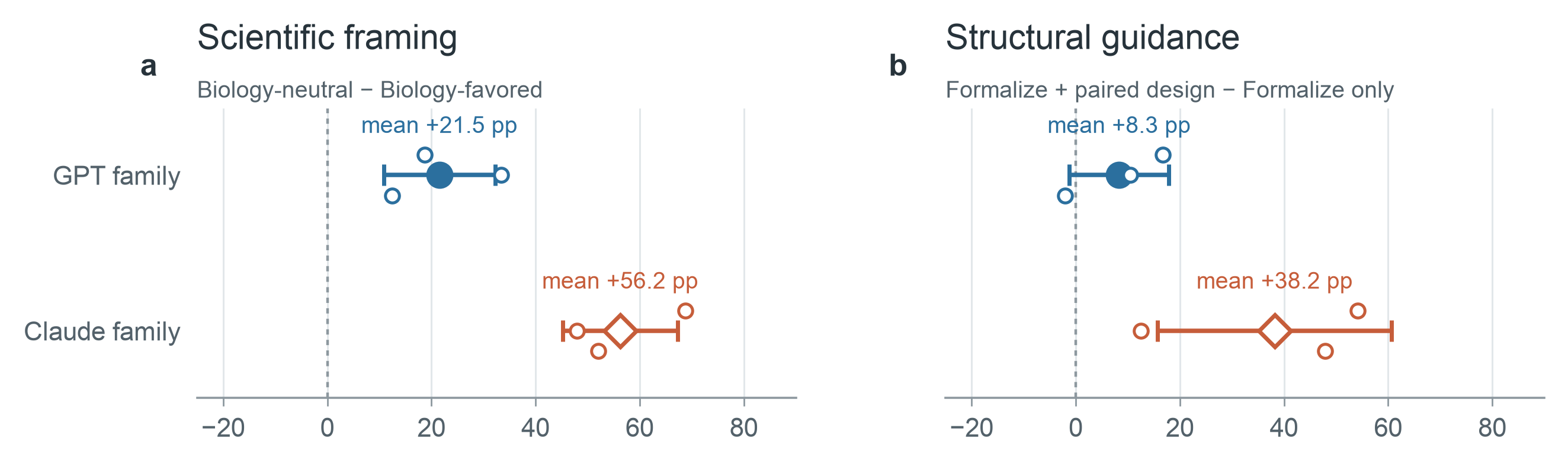}
  \caption{\textbf{Direct comparison of manipulation effects across the tested model variants.}
  Panel (a) shows the paired change from Biology-favored to Biology-neutral; panel (b) shows the paired change from Formalize only to Formalize + paired design. Small points are the three model-specific changes within each family. Large symbols show the mean across the three tested variants in each family, and horizontal bars show $\pm1$ standard deviation across those model-level changes. GPT includes GPT-5.6 Sol, GPT-5.6 Terra, and GPT-5.5; Claude includes Claude Opus 5, Claude Opus 4.8, and Claude Sonnet 5. Values use all 48 scheduled calls per condition, with invalid outputs and timeouts counted as failures.}
  \label{fig:additional-models}
\end{figure}

\begin{table}[p]
\centering
\caption{\textbf{Exact outcome counts across all evaluated models and
conditions.} The 23/24 and 22/24 columns report exact recovery of the
constraint-maximizing assignment and the prespecified nearby alternative,
respectively. ``Other'' includes every different assignment, invalid output,
or timeout. Each condition contains 48 scheduled calls.}
\label{tab:all-model-outcomes}
\small
\setlength{\tabcolsep}{5pt}
\begin{tabular}{@{}llrrrr@{}}
\toprule
Model & Condition & $n$ & 23/24 & 22/24 & Other \\
\midrule

\multirow{5}{*}{GPT-5.6 Sol}
  & Formal                    & 48 & 43 & 3  & 2 \\
  & Biology-favored          & 48 & 31 & 15 & 2 \\
  & Biology-neutral          & 48 & 37 & 9  & 2 \\
  & Formalize only           & 48 & 42 & 5  & 1 \\
  & Formalize + paired design & 48 & 41 & 4  & 3 \\
\addlinespace

\multirow{5}{*}{GPT-5.6 Terra}
  & Formal                    & 48 & 35 & 8  & 5 \\
  & Biology-favored          & 48 & 13 & 34 & 1 \\
  & Biology-neutral          & 48 & 29 & 14 & 5 \\
  & Formalize only           & 48 & 24 & 17 & 7 \\
  & Formalize + paired design & 48 & 29 & 15 & 4 \\
\addlinespace

\multirow{5}{*}{GPT-5.5}
  & Formal                    & 48 & 35 & 6  & 7 \\
  & Biology-favored          & 48 & 6  & 25 & 17 \\
  & Biology-neutral          & 48 & 15 & 3  & 30 \\
  & Formalize only           & 48 & 15 & 18 & 15 \\
  & Formalize + paired design & 48 & 23 & 9  & 16 \\
\addlinespace

\multirow{5}{*}{Claude Opus 5}
  & Formal                    & 48 & 46 & 1  & 1 \\
  & Biology-favored          & 48 & 13 & 34 & 1 \\
  & Biology-neutral          & 48 & 38 & 1  & 9 \\
  & Formalize only           & 48 & 21 & 27 & 0 \\
  & Formalize + paired design & 48 & 44 & 4  & 0 \\
\addlinespace

\multirow{5}{*}{Claude Opus 4.8}
  & Formal                    & 48 & 48 & 0  & 0 \\
  & Biology-favored          & 48 & 15 & 16 & 17 \\
  & Biology-neutral          & 48 & 38 & 1  & 9 \\
  & Formalize only           & 48 & 27 & 8  & 13 \\
  & Formalize + paired design & 48 & 33 & 3  & 12 \\
\addlinespace

\multirow{5}{*}{Claude Sonnet 5}
  & Formal                    & 48 & 44 & 4  & 0 \\
  & Biology-favored          & 48 & 0  & 42 & 6 \\
  & Biology-neutral          & 48 & 33 & 4  & 11 \\
  & Formalize only           & 48 & 4  & 37 & 7 \\
  & Formalize + paired design & 48 & 30 & 15 & 3 \\
\bottomrule
\end{tabular}
\end{table}

\paragraph{GPT-5.6 Terra.}
GPT-5.6 Terra recovered the 23/24 assignment in 73\% of Formal cases.
Recovery increased from 27\% in Biology-favored to 60\% in Biology-neutral,
and from 50\% under Formalize only to 60\% under Formalize + paired design.
Its formal baseline is lower than GPT-5.6 Sol's and below 75\%, so we interpret these manipulation effects
descriptively.

\paragraph{GPT-5.5.}
GPT-5.5 recovered the target in 73\% of Formal cases. Target recovery increased from
13\% in Biology-favored to 31\% in Biology-neutral, and from 31\% under the
generic formalization request to 48\% under Formalize + paired design.
Because the model did not pass the formal baseline gate, these changes should
not be interpreted as evidence that it robustly preserved the intended
constraint system in prose.

\paragraph{Claude Opus 4.8.}
Claude Opus 4.8 recovered the 23/24 assignment in 100\% of Formal cases.
Recovery increased from 31\% in Biology-favored to 79\% in Biology-neutral.
Unlike the other two Claude variants, structural guidance produced only a
modest increase in exact 23/24 recovery, from 56\% under Formalize only to
69\% under Formalize + paired design.

\paragraph{Claude Sonnet 5.}
Claude Sonnet 5 recovered the target in 92\% of Formal cases. Its responses
showed the same qualitative pattern as Claude Opus 5. In Biology-favored,
0\% of calls recovered the target and 88\% selected the exact 22/24
alternative. In Biology-neutral, target recovery rose to 69\%. Under the
generic formalization request, 8\% of calls recovered the target and 77\%
selected the alternative; under Formalize + paired design, target recovery
increased to 63\%. Both paired changes were positive across every
prespecified graph topology and prose family.

\paragraph{Interpretation.} Across the three tested variants from each family, the Claude variants show larger scientific-framing effects than the GPT variants. The paired-design effect is also larger on average for Claude, although it varies substantially across the three Claude models. We treat these summaries as descriptive patterns across the selected models rather than estimates of a broader model-family effect. Model family is confounded with differences in training, inference procedure, and capability, and GPT-5.5 and GPT-5.6 Terra differ from GPT-5.6 Sol in their formal baseline performance.

\section{Efficiency analysis}

\begin{figure}[H]
  \centering
  \includegraphics[width=0.94\textwidth]{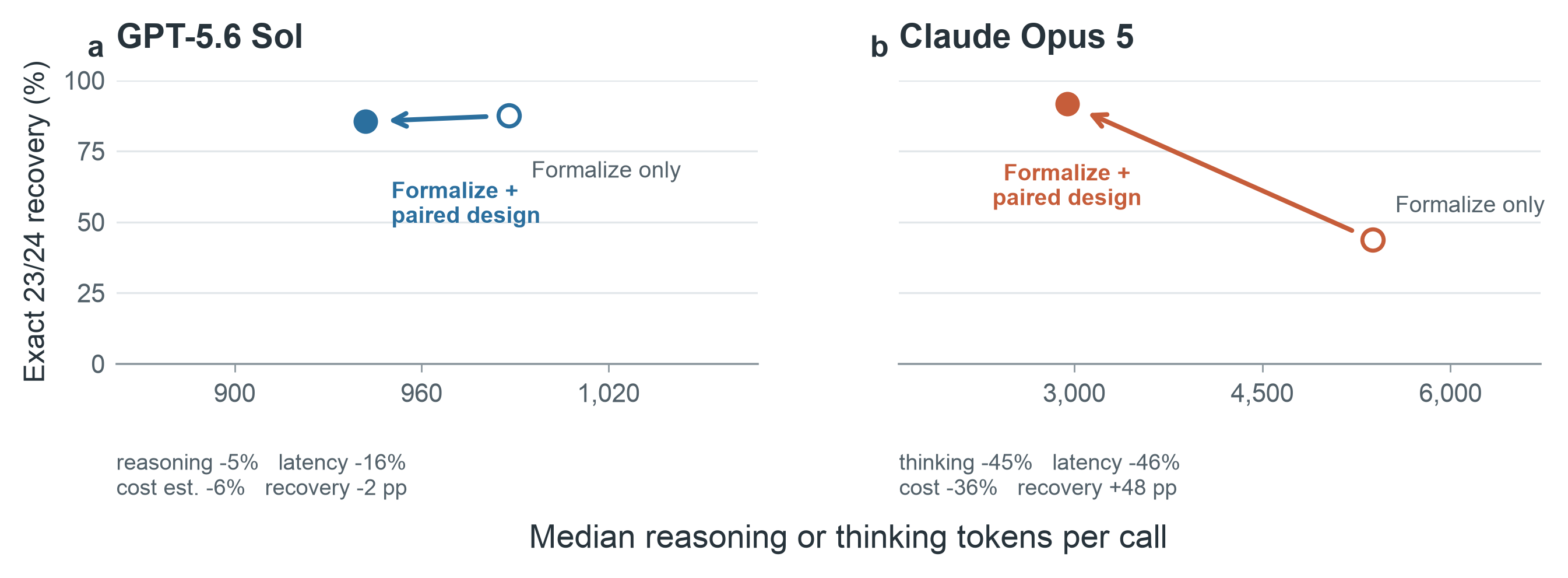}
\caption{\textbf{Descriptive efficiency comparison for the primary models.} Each arrow goes from
Formalize only to Formalize + paired design. Tokens and latency are medians. GPT
cost is an API-equivalent estimate from logged usage; Claude cost is
provider-reported. Reasoning-token scales are model-specific and should not be
compared across panels.}
  \label{fig:efficiency}
\end{figure}

For the two primary models, the paired-design sentence reduces median reasoning
tokens, latency, and cost. These descriptive efficiency comparisons do not
establish how either model used the added sentence, and accuracy differs
between the two conditions, especially for Claude.

\section{Limitations and scope}

This study is deliberately narrow. The scientific records are synthetic and do
not capture the full ambiguity of real scientific literature. The expected
biological directions are model-proposed synthetic priors rather than
independently validated biological facts. The primary evaluation covers two
models, 48 cases, and one response for each model--condition--case combination;
Appendix~\ref{app:additional-models} reports the same evaluation for four
additional model variants, giving three tested variants from each model family.
The formal problem is also restricted to a binary XOR setting with one
intentionally inconsistent relation.

Our measurements are limited to the models' final answers. We do not directly
observe the pairings they infer, the constraint systems they construct, or the
internal representations used to reach their conclusions. The results therefore
identify sensitivity near the boundary between scientific interpretation and
formal reasoning, but do not identify the representation responsible for it.

\end{document}